\pdfoutput=1
\documentclass[11pt]{article}

\usepackage[]{EMNLP2023}

\usepackage{times}
\usepackage{latexsym}
\usepackage{graphicx} 
\usepackage{amsmath}

\usepackage[T1]{fontenc}
\usepackage[utf8]{inputenc}

\usepackage{microtype}

\usepackage{inconsolata}

\title{RST-style Discourse Parsing Guided by Document-level Content Structures}

\author{Ming Li \\
  Texas A\&M University \\
  \texttt{liming@tamu.edu} \\\And
  Ruihong Huang \\
  Texas A\&M University \\
  \texttt{huangrh@cse.tamu.edu}}

\begin{document}
\maketitle
\begin{abstract}

Rhetorical Structure Theory based Discourse Parsing (RST-DP) explores how clauses, sentences, and large text spans compose a whole discourse and presents the rhetorical structure as a hierarchical tree. Existing RST parsing pipelines construct rhetorical structures without the knowledge of document-level content structures, which causes relatively low performance when predicting the discourse relations for large text spans. 
Recognizing the value of high-level content-related information in facilitating discourse relation recognition, we propose a novel pipeline for RST-DP that incorporates structure-aware news content sentence representations derived from the task of News Discourse Profiling. By incorporating only a few additional layers, this enhanced pipeline exhibits promising performance across various RST parsing metrics.

\end{abstract}

\section{Introduction}

\begin{figure}[tb]
\centering 
\includegraphics[width=0.48\textwidth]{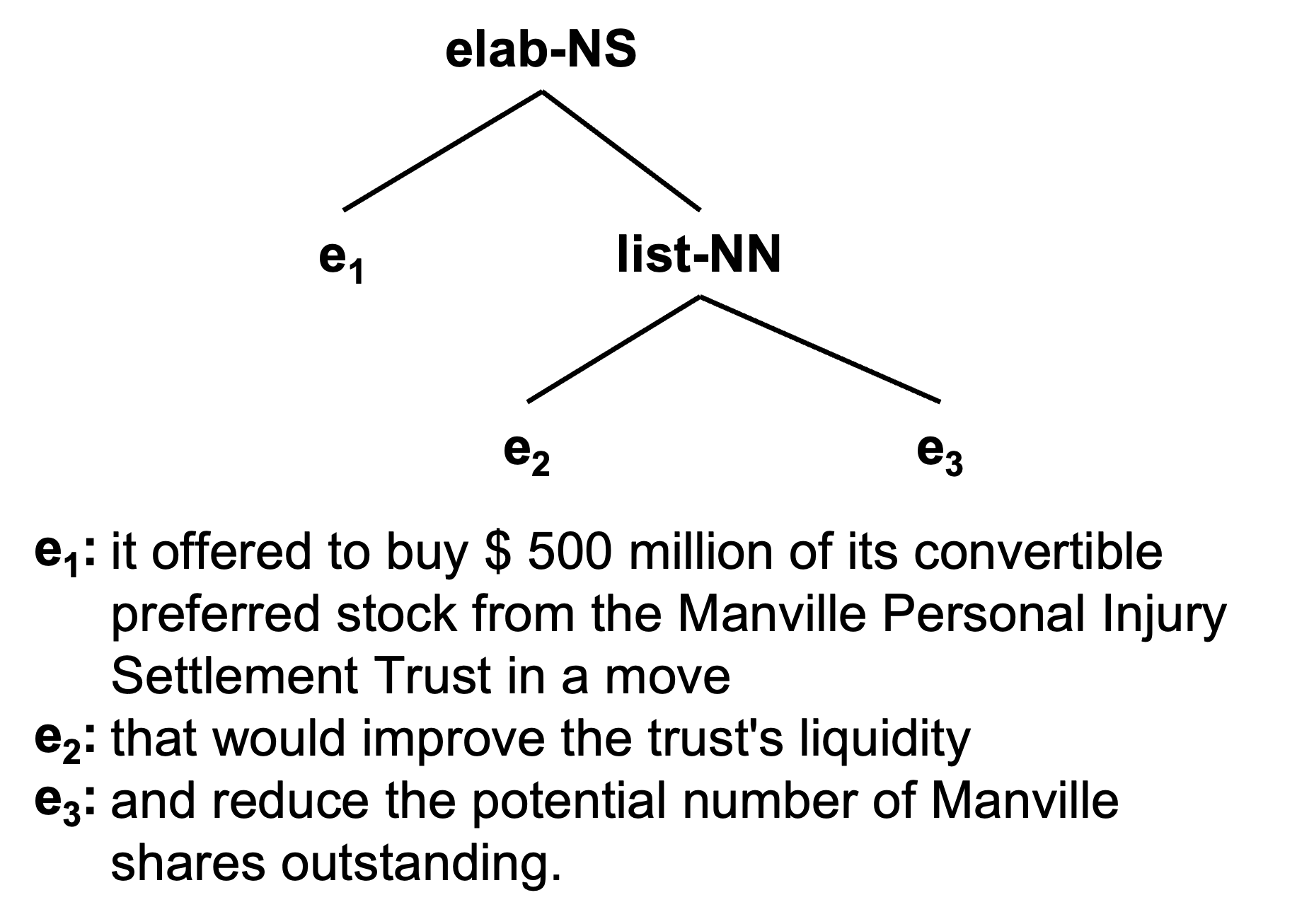} 
\caption{
An example discourse tree from English RST Discourse Treebank \cite{carlson-etal-2001-building} development set with small text spans. 
$e_1$, $e_2$ and $e_3$ are EDUs, \textit{elab} and \textit{list} are rhetorical relations. 
\textit{NS}, \textit{NN} are nuclearity relations where \textit{N} represents the nucleus and \textit{S} represents the satellite. 
} 
\label{intro} 
\end{figure}

Rhetorical Structure Theory based Discourse Parsing (RST-DP) \cite{mann1988rhetorical} aims to elucidate a hierarchical representation of the rhetorical structure within discourse by 
%employing a tree-based framework.
constructing a tree. 
Each leaf node in an RST tree represents an elementary discourse unit (EDU) while each internal node represents the relation between two text spans (an example shown in Figure \ref{intro}). 
It explores how clauses, sentences, and large text spans compose a whole discourse, which is useful for many NLP applications \cite{KRAUS201965, isonuma-etal-2019-unsupervised, spangher-etal-2021-multitask}. 

Early RST models \cite{joty-etal-2013-combining, li-etal-2014-recursive, li-etal-2016-discourse,wang-etal-2017-two} mostly utilize bottom-up approaches which limit the tree construction to consider only local context while recent models \cite{Kobayashi_Hirao_Kamigaito_Okumura_Nagata_2020, zhang-etal-2020-top, koto-etal-2021-top} mostly follow a top-down paradigm to fully utilize the global context. 
Despite the achievements of previous models, they might overlook the performance discrepancy in predicting rhetorical relations between small and large text spans. The failure to address this disparity resulted in an inadequate exploration of the potential benefits offered by document-level content structures in guiding the recognition of rhetorical relations.

\begin{figure*}[t]
\centering 
\includegraphics[width=1\textwidth]{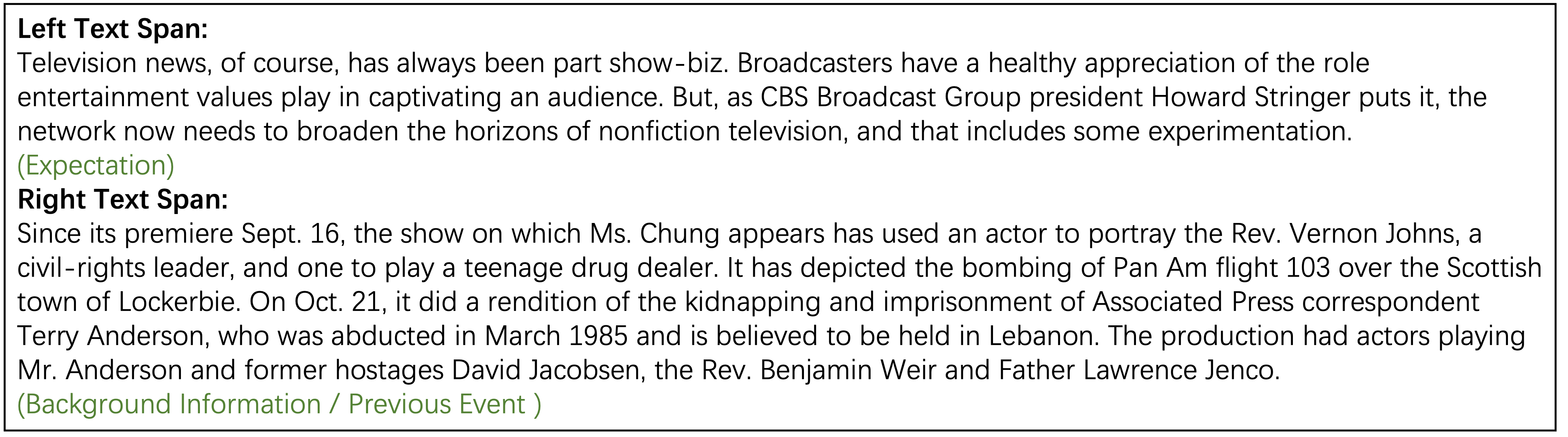} 
\caption{
An example with relatively large text spans. 
The model needs to predict the discourse relation and nuclear relation between these two large text spans. 
The golden label of this example is \textit{Elaboration} and \textit{Nuclear-Satellite}. 
In parentheses (green) are the possible news content type labels predicted by the news discourse profiling system. 
The incorporation of news content type labels serves as a valuable aid in comprehending the core information encapsulated within lengthy textual contexts, thereby facilitating informed decision-making. 
}
\label{large_example} 
\end{figure*}

Since RST discourse parsing aims to present the whole document into one tree, there is a significant disparity in length and amounts between the smallest text spans and relatively larger text spans. 
% largest and smallest subtrees. 
The smallest text spans (EDUs) are at clause level with large amounts in one discourse while larger text spans can be paragraph-level with only a few in each discourse. 
This substantial difference in length and amounts poses challenges for existing methods in maintaining their performance when confronted with exceptionally lengthy text spans.\footnote{Quantitive analysis on the performance disparity between small and large text spans can be found in Appendix \ref{sec:appendix_textspan}.}
To accurately identify the discourse relation between two relatively large text spans, the model must possess a comprehensive understanding of the content contained within both segments which is a nontrivial task. 
% Moreover, due to the composition of large text spans from smaller text spans, the training data exhibit a severely imbalanced distribution. Consequently, this skewed distribution inherently leads the model to overlook the larger spans. 

Based on the above discussion, the primary objective of this paper is to integrate the news-specific content information into the framework of RST discourse parsing as extra guidance on recognizing the relations between text spans. 
Specifically, the content information is derived from the news discourse profiling (NDP) task, which assigns one of eight distinct content types to each sentence within a news article. 
The content types include the main news event description, its immediate consequences, its direct context informing contents, and other supporting contents. \footnote{NDP task is introduced in detail in Appendix \ref{sec:appendix_ndp}.} 
These content types effectively capture the common discourse roles of sentences when describing a news story, which requires a comprehensive understanding of the underlying document-level content structures. 
This document-level content can serve as valuable guidance for the classification of nuclearity and rhetorical relations, especially for large text spans. 

Figure \ref{intro} presents an illustrative discourse tree from the English RST Discourse dataset \cite{carlson-etal-2001-building}. This particular discourse tree comprises three EDUs. 
\textit{elab} and \textit{list} are rhetorical relations and \textit{NS}, \textit{NN} are nuclearity relations. Given their simplicity, RST models can easily comprehend the primary content of each EDU and proceed to analyze their interrelations. However, complexities arise when the textual spans escalate in size. 
Figure \ref{large_example} showcases an instance involving relatively large text spans. Unlike the recognition of small spans, accurately predicting large spans necessitates a profound comprehension of the textual content. 
With the provided discourse-level content information in parentheses\footnote{Labels in parentheses are used to illustrate how content information help grasp the key points of text spans. The final predicted categories are not needed in our system. }, it reduces the overload for the model itself to extract its main content but can concentrate more on relation analysis. 
The incorporation of this content-type information serves as a valuable aid in comprehending the core content encapsulated within lengthy textual contexts, thereby facilitating informed decision-making.

Therefore, we incorporate content structures and introduce a new pipeline for RST discourse parsing called C2RNet, (\textbf{\underline C}2\textbf{{\underline R}} is short for ''from \textbf{\underline C}ontent structure to \textbf{{\underline R}}hetorical structure''), where two learning branches conduct news discourse profiling (NDP branch) and RST discourse parsing (RST branch) respectively and sentence embeddings derived from the NDP branch will be sent to the RST branch for RST discourse parsing.  
The two branches operate in parallel, ensuring that additional processing time is not required. To our knowledge, we are the first to incorporate extra news content structure to facilitate RST parsing.  Comparative evaluations against a baseline system that does not leverage content structures demonstrate the promising performance of C2RNet on the English RST Discourse dataset.

\section{The Architecture of C2RNet}

\begin{figure}[tb]
\centering 
\includegraphics[width=0.5\textwidth]{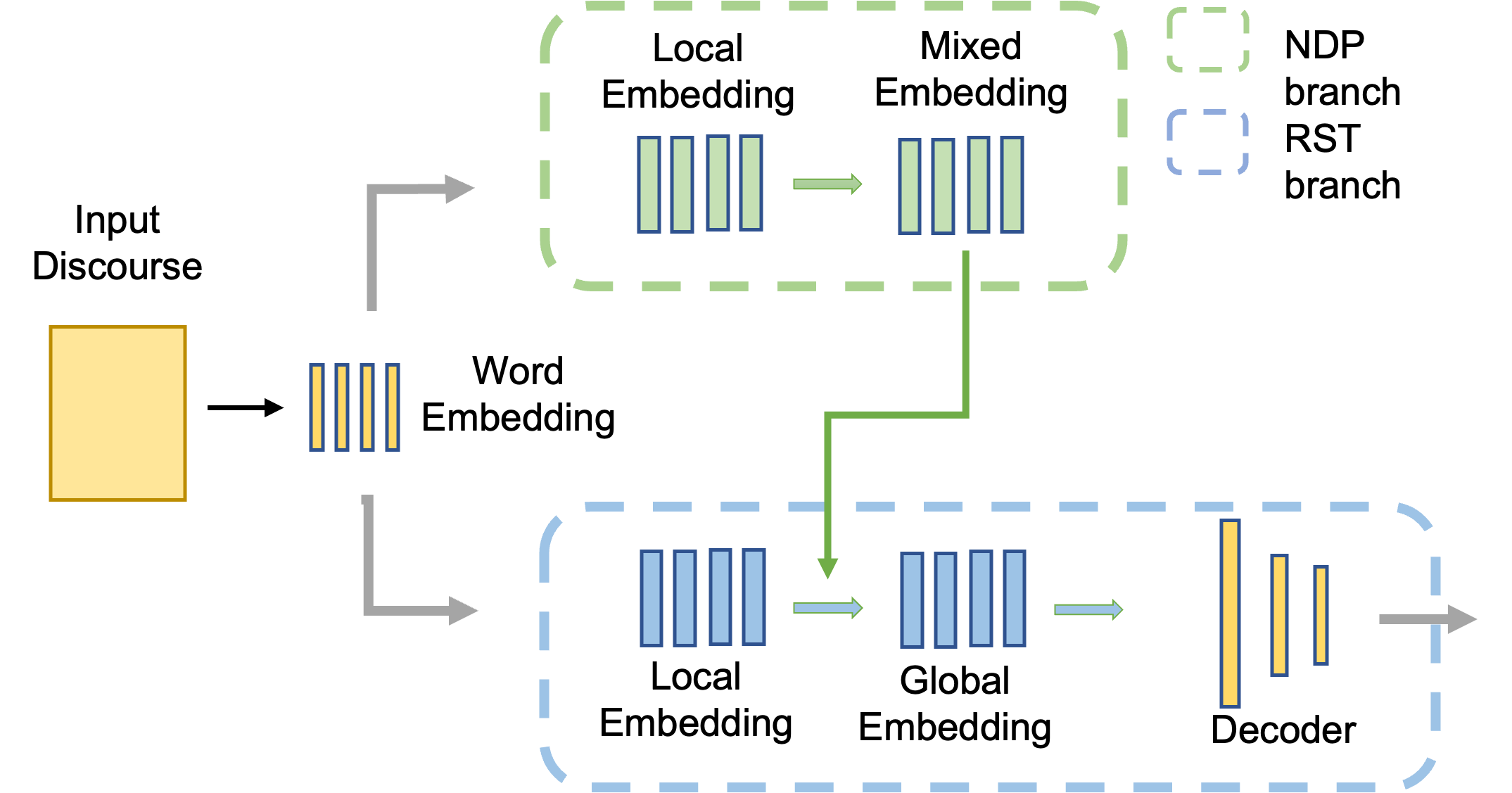} 
\caption{
Overall C2RNet pipeline. 
The NDP branch is represented by the upper dashed box, while the RST branch is depicted by the lower dashed box. Initially, the input discourse is fed into a pretrained language model to generate shared word embeddings. Subsequently, both the NDP branch and RST branch operate in parallel. The combined sentence embedding derived from the NDP branch is then transmitted to the RST branch, enabling the incorporation of comprehensive event-related knowledge.
} 
\label{pipeline} 
\end{figure}

The pipeline illustrated in Figure \ref{pipeline} comprises three key components: a shared language model, an NDP branch, and an RST branch. Given that our primary objective is to incorporate content-type information into the RST discourse parsing task, instead of designing intricate model architectures, we adopt existing off-the-shelf model structures to validate the effectiveness of our proposed method.

\subsection{NDP branch}

To maintain the overall efficiency of the pipeline, we adopt the NDP branch proposed by LimNet \cite{limnet}. This branch incorporates two self-attention modules \cite{bahdanau2014neural, NIPS2015_1068c6e4} to generate embeddings at the EDU level.
The initial self-attention module computes attention weights for each word within the local EDU, resulting in local EDU embeddings. Subsequently, the second self-attention module calculates attention weights across the entire discourse, yielding the global EDU embeddings. Finally, the mixed EDU embeddings are obtained by combining the local and global EDU embeddings through addition.

\subsection{RST branch}

The RST branch in our approach is based on the hierarchical structure proposed by \cite{koto-etal-2021-top}. We re-implement this structure, which consists of two BiLSTM layers, to capture the discourse structure. The first BiLSTM layer is applied to the local Elementary Discourse Units (EDUs), followed by average pooling to obtain the local EDU embeddings. These local EDU embeddings are then combined with the mixed EDU embeddings obtained from the NDP branch. The concatenated embeddings serve as the input for the second BiLSTM layer, which produces the global EDU embeddings for the RST segmenter. 

For the RST decoder part, we follow the same implementation from \citet{koto-etal-2021-top}, where implicit paragraph boundary features are concatenated and the RST trees are constructed iteratively. 
During each iteration, the input sentence is split at the position with the highest probability, which is determined by the model. To obtain representative vectors for the left and right text spans at each splitting iteration, average pooling operations are applied to the corresponding vectors. These left and right vectors are then concatenated together and passed through a final fully-connected layer. This final layer generates the joint probability distribution over the nuclearity and rhetorical relations.

\section{Experiments}

\subsection{Dataset}

% \vspace{.05in}
% \noindent
% \textbf{RST data}

We use the English RST Discourse Treebank \cite{carlson-etal-2001-building} for our C2RNet. 
It is based on the Wall Street Journal portion of the Penn Treebank \cite{marcus-etal-1993-building}, which contains $347$ documents for training, and $38$ documents for testing. 
For a fair comparison, we use the same development set as \citet{koto-etal-2021-top} which contains $35$ documents from the training set.

\begin{table*}[h]
\centering
\scalebox{0.9}{\begin{tabular}{l|c|c|c|c|c|c|c|c}
\hline
& \multicolumn{4}{|c|}{\textbf{Original Parseval}} & \multicolumn{4}{|c}{\textbf{RST Parseval}}\\ 
\hline
\ & \textbf{S} & \textbf{N} & \textbf{R} & \textbf{F} & \textbf{S} & \textbf{N} & \textbf{R} & \textbf{F}  \\ \hline
\ \cite{hayashi-etal-2016-empirical} & $65.1$ & $54.6$ & $44.7$ & $44.1$
& $ 82.6 $ & $ 66.6 $ & $ 54.6 $ & $ 54.3 $  \\
\ \cite{li-etal-2016-discourse} & $64.5$ & $54.0$ & $38.1$ & $36.6$
& $ 82.2 $ & $ 66.5 $ & $ 51.4 $ & $ 50.6 $   \\
\ \cite{braud-etal-2017-cross} & $62.7$ & $54.5$ & $45.5$ & $45.1$
& $ 81.3 $ & $ 68.1 $ & $ 56.3 $ & $ 56.0 $   \\
\ \cite{yu-etal-2018-transition} & $71.4$ & $60.3$ & $49.2$ & $48.1$
& $ 85.6 $ & $ 72.9 $ & $ 59.8 $ & $ 59.3 $    \\
\ \cite{mabona-etal-2019-neural} & $67.1$ & $57.4$ & $45.5$ & $45.0$ 
& - & - & - & -   \\
\ \cite{Kobayashi_Hirao_Kamigaito_Okumura_Nagata_2020} & - & - & - & - 
& $ 87.0 $ & $ 74.6 $ & $ 60.0 $ & -  \\
\ \cite{zhang-etal-2020-top} & $ 67.2 $ & $ 55.5 $ & $ 45.3 $ & $ 44.3 $ 
& - & - & - & -  \\
\ \cite{nguyen-etal-2021-rst} & $ 74.3 $ & $ 64.3 $ & $ 51.6 $ & $ 50.2 $ 
& $87.6$ & $76.0$ & $61.8$  \\
\ \cite{koto-etal-2021-top} & $ 73.1 $ & {$ 62.3 $} & $ 51.5 $ & $ 50.3 $ 
& {$86.6 $} & $73.7 $ & $ 61.5 $ & $ 60.9 $  \\
\ \cite{zhang-etal-2021-adversarial} & $ 76.3 $ & {$ 65.5 $} & $ \bf 55.6 $ & $ \bf 53.8 $ 
& - & - & - & -  \\
\ \cite{yu-etal-2022-rst} & $ 76.4 $ & {$ 66.1 $} & $ 54.5 $ & $ 53.5 $ 
& - & - & - & -  \\
\hline 
\ Baseline & $ 75.4 $ & $ 64.1 $ & $ 53.6 $ & $ 52.1$ 
& $ 87.7 $ & $ 75.0 $ & $ 63.1 $ & $ 62.3 $  \\
\ C2RNet (ours) & $ \bf 76.8 $ & $ \bf 66.2 $ & {$  55.4 $} & {$ \bf 53.8$ }
& {$ \bf 88.4 $} & {$ \bf 76.5 $} & {$ \bf 64.5 $} & {$ \bf 63.8 $}  \\ \hline

\ Human & $ 78.7 $ & $ 66.8 $ & $ 57.1 $ & $ 55.0 $ 
& $ 88.3 $ & $ 77.3 $ & $ 65.4 $ & $ 64.7 $ \\
\hline
\end{tabular}}
\caption{\label{tbl:rst}
RST discourse parsing results on the test set of the RST dataset, using original Parseval and RST Parseval metrics \cite{marcu-2000-rhetorical}. 
S, N, R, and F represent Span, Nuclearity, Relation, and Full. 
The results of ours are averaged over $3$ random runs and the highest performances are in bold. 
\textit{Baseline} represents the model without an NDP branch and other configurations are kept the same. 
}
\end{table*} % compared with sota

\subsection{Implementation Details}

The NDP model LiMNet \cite{limnet} was trained in advance and the weights are transferred to the NDP branch of C2RNet as the initial weights. The weights of the NDP branch are fixed for the initial $40$ epochs and then optimized under only RST data. 
Our pipeline is trained using Adam optimizer \cite{kingma2014adam} with the initial learning rate of $5e-4$, epsilon of $1e-6$ for $150$ epochs. 
The dropout rate \cite{JMLR:v15:srivastava14a} is $0.5$ for all experiments. 
 By default, we employ the \emph{t5-large} model \cite{JMLR:v21:20-074} from the \emph{huggingface} library \cite{wolf2019huggingface} as our pretrained language model. The pretrained language model is not fine-tuned but rather kept fixed to ensure the efficiency and scalability of our model.
\footnote{The performance of our model with other language models is provided in Appendix \ref{sec:appendix_lm}, where we present the results and analyses for different language model configurations.}

\subsection{Evaluation}

Table \ref{tbl:rst} presents the results of various models, examining their performance in RST discourse parsing. The \textit{Baseline} model refers to the configuration where the NDP branch is excluded, while maintaining all other settings identical to C2RNet, including the pretrained language model. Our \textit{C2RNet} model consistently surpasses the \textit{Baseline} across all metrics. These findings affirm that incorporating content structure-aware sentence representations provides notable advantages in all aspects of RST discourse parsing, encompassing span identification, nuclearity prediction, and rhetorical relation recognition.
\footnote{Detailed analysis on the effects of NDP labels is presented in Appendix \ref{sec:appendix_ndpl}.}

\subsection{Result Analysis}

We compared the performance of our model and the baseline model on spans of different lengths. The length of a span is defined as the number of EDUs subsumed by a subtree. We divided the spans into three groups based on their lengths. The first group consists of minimal length spans containing exactly $2$ EDUs, (for example, $e_2$ and $e_3$ in Figure \ref{intro}), which accounts for $34.4\%$ of all data. The remaining spans were split into medium-length spans ($3$, $4$, and $5$ EDUs) and long spans (more than $5$ EDUs), (for example, two text spans in Figure \ref{large_example}), representing $35.5\%$ and $30\%$ of all spans, respectively. 
Table \ref{tbl:ablation_rf} displays the accuracy gaps between our model and the baseline model, depicting the variances when predicting different text spans within the test set. The results reveal that the incorporation of content structures proves to be particularly advantageous in the parsing of longer text spans, as observed in both nuclearity and rhetorical relation prediction. Specifically, our model achieves superior performance in nuclearity prediction for both medium-length and very long spans, while for rhetorical relation prediction, our model demonstrates a significant advantage primarily in long spans. \footnote{Detailed table and analysis is provided in Appendix \ref{sec:appendix_textspan}.}

\begin{table}[t]
\centering
\scalebox{0.75}{\begin{tabular}{l|c|c|c}
\hline
\ & Span $= 2$ & $2 < $ Span $\leq 5$ & Span $>5$  \\ \hline
\  Nuclearity  & $ -0.4 \%$ & $ +2.0 \%$ & $ +1.9 \%$ \\ 
\  Rhetorical relations & $ 0 \%$ & $ -0.7 \%$ & $ +1.6 \%$ \\
\hline
\end{tabular}}
\caption{\label{tbl:ablation_rf}
The accuracy gaps between our model and the baseline model, when parsing text spans of different lengths in the test set. 
}
\end{table}

\section{Conclusion}

In this paper, we propose a novel pipeline C2RNet for RST discourse parsing that leverages document-level content knowledge to enhance the recognition of rhetorical relations. Our approach incorporates content structure-aware sentence representations, which prove to be beneficial for enhancing discourse parsing, encompassing span identification, nuclearity prediction, and rhetorical relation prediction, especially for relatively large text spans. With only a few layers added, our model achieves promising performance. 

\section*{Limitations}

A key limitation of our proposed approach lies in its dependence on advanced language models. In order to maintain efficiency throughout the pipeline, we employ a single shared pretrained language model for both branches, necessitating the selection of a sufficiently powerful language model capable of supporting both tasks. Our additional results, presented in the appendix, demonstrate that employing a weaker language model leads to a decrease in performance. Nonetheless, our C2RNet model consistently outperforms the baseline models by a considerable margin, highlighting its robustness and effectiveness.

% Entries for the entire Anthology, followed by custom entries
\bibliography{anthology,custom}

\begin{thebibliography}{32}
\expandafter\ifx\csname natexlab\endcsname\relax\def\natexlab#1{#1}\fi

\bibitem[{Bahdanau et~al.(2014)Bahdanau, Cho, and Bengio}]{bahdanau2014neural}
Dzmitry Bahdanau, Kyunghyun Cho, and Yoshua Bengio. 2014.
\newblock Neural machine translation by jointly learning to align and translate.
\newblock \emph{arXiv preprint arXiv:1409.0473}.

\bibitem[{Braud et~al.(2017)Braud, Coavoux, and S{\o}gaard}]{braud-etal-2017-cross}
Chlo{\'e} Braud, Maximin Coavoux, and Anders S{\o}gaard. 2017.
\newblock \href {https://aclanthology.org/E17-1028} {Cross-lingual {RST} discourse parsing}.
\newblock In \emph{Proceedings of the 15th Conference of the {E}uropean Chapter of the Association for Computational Linguistics: Volume 1, Long Papers}, pages 292--304, Valencia, Spain. Association for Computational Linguistics.

\bibitem[{Carlson et~al.(2001)Carlson, Marcu, and Okurovsky}]{carlson-etal-2001-building}
Lynn Carlson, Daniel Marcu, and Mary~Ellen Okurovsky. 2001.
\newblock \href {https://aclanthology.org/W01-1605} {Building a discourse-tagged corpus in the framework of {R}hetorical {S}tructure {T}heory}.
\newblock In \emph{Proceedings of the Second {SIG}dial Workshop on Discourse and Dialogue}.

\bibitem[{Chorowski et~al.(2015)Chorowski, Bahdanau, Serdyuk, Cho, and Bengio}]{NIPS2015_1068c6e4}
Jan~K Chorowski, Dzmitry Bahdanau, Dmitriy Serdyuk, Kyunghyun Cho, and Yoshua Bengio. 2015.
\newblock \href {https://proceedings.neurips.cc/paper/2015/file/1068c6e4c8051cfd4e9ea8072e3189e2-Paper.pdf} {Attention-based models for speech recognition}.
\newblock In \emph{Advances in Neural Information Processing Systems}, volume~28. Curran Associates, Inc.

\bibitem[{Choubey and Huang(2021)}]{choubey-huang-2021-profiling-news}
Prafulla~Kumar Choubey and Ruihong Huang. 2021.
\newblock \href {https://doi.org/10.18653/v1/2021.findings-emnlp.137} {{P}rofiling news discourse structure using explicit subtopic structures guided critics}.
\newblock In \emph{Findings of the Association for Computational Linguistics: EMNLP 2021}, pages 1594--1605, Punta Cana, Dominican Republic. Association for Computational Linguistics.

\bibitem[{Choubey et~al.(2020)Choubey, Lee, Huang, and Wang}]{choubey-etal-2020-discourse}
Prafulla~Kumar Choubey, Aaron Lee, Ruihong Huang, and Lu~Wang. 2020.
\newblock \href {https://doi.org/10.18653/v1/2020.acl-main.478} {Discourse as a function of event: Profiling discourse structure in news articles around the main event}.
\newblock In \emph{Proceedings of the 58th Annual Meeting of the Association for Computational Linguistics}, pages 5374--5386, Online. Association for Computational Linguistics.

\bibitem[{Devlin et~al.(2019)Devlin, Chang, Lee, and Toutanova}]{devlin-etal-2019-bert}
Jacob Devlin, Ming-Wei Chang, Kenton Lee, and Kristina Toutanova. 2019.
\newblock \href {https://doi.org/10.18653/v1/N19-1423} {{BERT}: Pre-training of deep bidirectional transformers for language understanding}.
\newblock In \emph{Proceedings of the 2019 Conference of the North {A}merican Chapter of the Association for Computational Linguistics: Human Language Technologies, Volume 1 (Long and Short Papers)}, pages 4171--4186, Minneapolis, Minnesota. Association for Computational Linguistics.

\bibitem[{Hayashi et~al.(2016)Hayashi, Hirao, and Nagata}]{hayashi-etal-2016-empirical}
Katsuhiko Hayashi, Tsutomu Hirao, and Masaaki Nagata. 2016.
\newblock \href {https://doi.org/10.18653/v1/W16-3616} {Empirical comparison of dependency conversions for {RST} discourse trees}.
\newblock In \emph{Proceedings of the 17th Annual Meeting of the Special Interest Group on Discourse and Dialogue}, pages 128--136, Los Angeles. Association for Computational Linguistics.

\bibitem[{Isonuma et~al.(2019)Isonuma, Mori, and Sakata}]{isonuma-etal-2019-unsupervised}
Masaru Isonuma, Junichiro Mori, and Ichiro Sakata. 2019.
\newblock \href {https://doi.org/10.18653/v1/P19-1206} {Unsupervised neural single-document summarization of reviews via learning latent discourse structure and its ranking}.
\newblock In \emph{Proceedings of the 57th Annual Meeting of the Association for Computational Linguistics}, pages 2142--2152, Florence, Italy. Association for Computational Linguistics.

\bibitem[{Joty et~al.(2013)Joty, Carenini, Ng, and Mehdad}]{joty-etal-2013-combining}
Shafiq Joty, Giuseppe Carenini, Raymond Ng, and Yashar Mehdad. 2013.
\newblock \href {https://aclanthology.org/P13-1048} {Combining intra- and multi-sentential rhetorical parsing for document-level discourse analysis}.
\newblock In \emph{Proceedings of the 51st Annual Meeting of the Association for Computational Linguistics (Volume 1: Long Papers)}, pages 486--496, Sofia, Bulgaria. Association for Computational Linguistics.

\bibitem[{Kingma and Ba(2014)}]{kingma2014adam}
Diederik~P Kingma and Jimmy Ba. 2014.
\newblock Adam: A method for stochastic optimization.
\newblock \emph{arXiv preprint arXiv:1412.6980}.

\bibitem[{Kobayashi et~al.(2020)Kobayashi, Hirao, Kamigaito, Okumura, and Nagata}]{Kobayashi_Hirao_Kamigaito_Okumura_Nagata_2020}
Naoki Kobayashi, Tsutomu Hirao, Hidetaka Kamigaito, Manabu Okumura, and Masaaki Nagata. 2020.
\newblock \href {https://doi.org/10.1609/aaai.v34i05.6321} {Top-down rst parsing utilizing granularity levels in documents}.
\newblock \emph{Proceedings of the AAAI Conference on Artificial Intelligence}, 34(05):8099--8106.

\bibitem[{Koto et~al.(2021)Koto, Lau, and Baldwin}]{koto-etal-2021-top}
Fajri Koto, Jey~Han Lau, and Timothy Baldwin. 2021.
\newblock \href {https://doi.org/10.18653/v1/2021.eacl-main.60} {Top-down discourse parsing via sequence labelling}.
\newblock In \emph{Proceedings of the 16th Conference of the European Chapter of the Association for Computational Linguistics: Main Volume}, pages 715--726, Online. Association for Computational Linguistics.

\bibitem[{Kraus and Feuerriegel(2019)}]{KRAUS201965}
Mathias Kraus and Stefan Feuerriegel. 2019.
\newblock \href {https://doi.org/https://doi.org/10.1016/j.eswa.2018.10.002} {Sentiment analysis based on rhetorical structure theory:learning deep neural networks from discourse trees}.
\newblock \emph{Expert Systems with Applications}, 118:65--79.

\bibitem[{Li et~al.(2014)Li, Li, and Hovy}]{li-etal-2014-recursive}
Jiwei Li, Rumeng Li, and Eduard Hovy. 2014.
\newblock \href {https://doi.org/10.3115/v1/D14-1220} {Recursive deep models for discourse parsing}.
\newblock In \emph{Proceedings of the 2014 Conference on Empirical Methods in Natural Language Processing ({EMNLP})}, pages 2061--2069, Doha, Qatar. Association for Computational Linguistics.

\bibitem[{Li et~al.(2022)Li, Yu, and Huang}]{limnet}
Ming Li, Sijing Yu, and Ruihong Huang. 2022.
\newblock Less is more: Simplifying feature extractors prevents overfitting for neural discourse parsing models.
\newblock \emph{arXiv preprint arXiv:2210.09537}.

\bibitem[{Li et~al.(2016)Li, Li, and Chang}]{li-etal-2016-discourse}
Qi~Li, Tianshi Li, and Baobao Chang. 2016.
\newblock \href {https://doi.org/10.18653/v1/D16-1035} {Discourse parsing with attention-based hierarchical neural networks}.
\newblock In \emph{Proceedings of the 2016 Conference on Empirical Methods in Natural Language Processing}, pages 362--371, Austin, Texas. Association for Computational Linguistics.

\bibitem[{Liu et~al.(2019)Liu, Ott, Goyal, Du, Joshi, Chen, Levy, Lewis, Zettlemoyer, and Stoyanov}]{Liu2019RoBERTaAR}
Yinhan Liu, Myle Ott, Naman Goyal, Jingfei Du, Mandar Joshi, Danqi Chen, Omer Levy, Mike Lewis, Luke Zettlemoyer, and Veselin Stoyanov. 2019.
\newblock Roberta: A robustly optimized bert pretraining approach.
\newblock \emph{ArXiv}, abs/1907.11692.

\bibitem[{Mabona et~al.(2019)Mabona, Rimell, Clark, and Vlachos}]{mabona-etal-2019-neural}
Amandla Mabona, Laura Rimell, Stephen Clark, and Andreas Vlachos. 2019.
\newblock \href {https://doi.org/10.18653/v1/D19-1233} {Neural generative rhetorical structure parsing}.
\newblock In \emph{Proceedings of the 2019 Conference on Empirical Methods in Natural Language Processing and the 9th International Joint Conference on Natural Language Processing (EMNLP-IJCNLP)}, pages 2284--2295, Hong Kong, China. Association for Computational Linguistics.

\bibitem[{Mann and Thompson(1988)}]{mann1988rhetorical}
William~C Mann and Sandra~A Thompson. 1988.
\newblock Rhetorical structure theory: Toward a functional theory of text organization.
\newblock \emph{Text-interdisciplinary Journal for the Study of Discourse}, 8(3):243--281.

\bibitem[{Marcu(2000)}]{marcu-2000-rhetorical}
Daniel Marcu. 2000.
\newblock \href {https://aclanthology.org/J00-3005} {The rhetorical parsing of unrestricted texts: a surface-based approach}.
\newblock \emph{Computational Linguistics}, 26(3):395--448.

\bibitem[{Marcus et~al.(1993)Marcus, Santorini, and Marcinkiewicz}]{marcus-etal-1993-building}
Mitchell~P. Marcus, Beatrice Santorini, and Mary~Ann Marcinkiewicz. 1993.
\newblock \href {https://aclanthology.org/J93-2004} {Building a large annotated corpus of {E}nglish: The {P}enn {T}reebank}.
\newblock \emph{Computational Linguistics}, 19(2):313--330.

\bibitem[{Nguyen et~al.(2021)Nguyen, Nguyen, Joty, and Li}]{nguyen-etal-2021-rst}
Thanh-Tung Nguyen, Xuan-Phi Nguyen, Shafiq Joty, and Xiaoli Li. 2021.
\newblock \href {https://doi.org/10.18653/v1/2021.naacl-main.128} {{RST} parsing from scratch}.
\newblock In \emph{Proceedings of the 2021 Conference of the North American Chapter of the Association for Computational Linguistics: Human Language Technologies}, pages 1613--1625, Online. Association for Computational Linguistics.

\bibitem[{Raffel et~al.(2020)Raffel, Shazeer, Roberts, Lee, Narang, Matena, Zhou, Li, and Liu}]{JMLR:v21:20-074}
Colin Raffel, Noam Shazeer, Adam Roberts, Katherine Lee, Sharan Narang, Michael Matena, Yanqi Zhou, Wei Li, and Peter~J. Liu. 2020.
\newblock \href {http://jmlr.org/papers/v21/20-074.html} {Exploring the limits of transfer learning with a unified text-to-text transformer}.
\newblock \emph{Journal of Machine Learning Research}, 21(140):1--67.

\bibitem[{Spangher et~al.(2021)Spangher, May, Shiang, and Deng}]{spangher-etal-2021-multitask}
Alexander Spangher, Jonathan May, Sz-Rung Shiang, and Lingjia Deng. 2021.
\newblock \href {https://doi.org/10.18653/v1/2021.emnlp-main.40} {Multitask semi-supervised learning for class-imbalanced discourse classification}.
\newblock In \emph{Proceedings of the 2021 Conference on Empirical Methods in Natural Language Processing}, pages 498--517, Online and Punta Cana, Dominican Republic. Association for Computational Linguistics.

\bibitem[{Srivastava et~al.(2014)Srivastava, Hinton, Krizhevsky, Sutskever, and Salakhutdinov}]{JMLR:v15:srivastava14a}
Nitish Srivastava, Geoffrey Hinton, Alex Krizhevsky, Ilya Sutskever, and Ruslan Salakhutdinov. 2014.
\newblock \href {http://jmlr.org/papers/v15/srivastava14a.html} {Dropout: A simple way to prevent neural networks from overfitting}.
\newblock \emph{Journal of Machine Learning Research}, 15(56):1929--1958.

\bibitem[{Wang et~al.(2017)Wang, Li, and Wang}]{wang-etal-2017-two}
Yizhong Wang, Sujian Li, and Houfeng Wang. 2017.
\newblock \href {https://doi.org/10.18653/v1/P17-2029} {A two-stage parsing method for text-level discourse analysis}.
\newblock In \emph{Proceedings of the 55th Annual Meeting of the Association for Computational Linguistics (Volume 2: Short Papers)}, pages 184--188, Vancouver, Canada. Association for Computational Linguistics.

\bibitem[{Wolf et~al.(2019)Wolf, Debut, Sanh, Chaumond, Delangue, Moi, Cistac, Rault, Louf, Funtowicz et~al.}]{wolf2019huggingface}
Thomas Wolf, Lysandre Debut, Victor Sanh, Julien Chaumond, Clement Delangue, Anthony Moi, Pierric Cistac, Tim Rault, R{\'e}mi Louf, Morgan Funtowicz, et~al. 2019.
\newblock Huggingface's transformers: State-of-the-art natural language processing.
\newblock \emph{arXiv preprint arXiv:1910.03771}.

\bibitem[{Yu et~al.(2018)Yu, Zhang, and Fu}]{yu-etal-2018-transition}
Nan Yu, Meishan Zhang, and Guohong Fu. 2018.
\newblock \href {https://aclanthology.org/C18-1047} {Transition-based neural {RST} parsing with implicit syntax features}.
\newblock In \emph{Proceedings of the 27th International Conference on Computational Linguistics}, pages 559--570, Santa Fe, New Mexico, USA. Association for Computational Linguistics.

\bibitem[{Yu et~al.(2022)Yu, Zhang, Fu, and Zhang}]{yu-etal-2022-rst}
Nan Yu, Meishan Zhang, Guohong Fu, and Min Zhang. 2022.
\newblock \href {https://doi.org/10.18653/v1/2022.acl-long.294} {{RST} discourse parsing with second-stage {EDU}-level pre-training}.
\newblock In \emph{Proceedings of the 60th Annual Meeting of the Association for Computational Linguistics (Volume 1: Long Papers)}, pages 4269--4280, Dublin, Ireland. Association for Computational Linguistics.

\bibitem[{Zhang et~al.(2021)Zhang, Kong, and Zhou}]{zhang-etal-2021-adversarial}
Longyin Zhang, Fang Kong, and Guodong Zhou. 2021.
\newblock \href {https://doi.org/10.18653/v1/2021.acl-long.305} {Adversarial learning for discourse rhetorical structure parsing}.
\newblock In \emph{Proceedings of the 59th Annual Meeting of the Association for Computational Linguistics and the 11th International Joint Conference on Natural Language Processing (Volume 1: Long Papers)}, pages 3946--3957, Online. Association for Computational Linguistics.

\bibitem[{Zhang et~al.(2020)Zhang, Xing, Kong, Li, and Zhou}]{zhang-etal-2020-top}
Longyin Zhang, Yuqing Xing, Fang Kong, Peifeng Li, and Guodong Zhou. 2020.
\newblock \href {https://doi.org/10.18653/v1/2020.acl-main.569} {A top-down neural architecture towards text-level parsing of discourse rhetorical structure}.
\newblock In \emph{Proceedings of the 58th Annual Meeting of the Association for Computational Linguistics}, pages 6386--6395, Online. Association for Computational Linguistics.

\end{thebibliography}
\bibliographystyle{acl_natbib}

\appendix
\section{Performance on different text spans}
\label{sec:appendix_textspan}

\begin{table*}[h]
\centering
\scalebox{0.9}{
\begin{tabular}{l|c|c|c|c|c|c|c|c|c|c|c}
\hline
Text Span Length  & 3 & 4 & 5 & 6 & 7 & 8 & 9 & 10 & 11 & 13 & 15 \\ \hline
Baseline ($>$ threshold ) & 43.3 & 37.9 & 33.3 & 30.8 & 30.5 & 30.2 & 28.1 & 27.2 & 26.8 & 27.3 & 26.9 \\
Baseline ($\leq$ threshold ) & 75.1 & 72.8 & 71.7 & 70.1 & 68.9 & 67.8 & 67.4 & 66.9 & 66.5 & 65.3 & 64.6 \\
C2RNet ($>$ threshold ) & 44.9 & 39.7 & 35.8 & 34.2 & 33.5 & 32.1 & 31.0 & 30.1 & 30.0 & 29.2 & 28.2 \\
C2RNet ($\leq$ threshold ) & 75.8 & 73.5 & 72.0 & 70.3 & 69.4 & 68.8 & 68.3 & 67.8 & 67.3 & 66.4 & 65.9 \\
\hline
Difference ($>$ threshold ) & 1.6 & 1.8 & 2.5 & 3.4 & 3.0 & 1.9 & 2.9 & 2.9 & 3.2 & 1.9 & 1.3 \\
Difference ($\leq$ threshold ) & 0.7 & 0.7 & 0.3 & 0.2 & 0.5 & 1.0 & 0.9 & 0.9 & 0.8 & 1.1 & 1.3 \\
\hline
\end{tabular}
}
\caption{\label{tbl:detail1}
Nuclear relation recognition accuracies in predicting text spans with different lengths. \textit{$>$ threshold} represents the overall accuracy when predicting the text spans greater than the given length. \textit{$\leq$ threshold} represents the overall accuracy when predicting the text spans less or equal to the given length. \textit{Difference} represents the gap between the baseline model and our C2RNet. 
}
\end{table*}

\begin{table*}[h]
\centering
\scalebox{0.9}{
\begin{tabular}{l|c|c|c|c|c|c|c|c|c|c|c}
\hline
Text Span Length  & 3 & 4 & 5 & 6 & 7 & 8 & 9 & 10 & 11 & 13 & 15 \\ \hline
Baseline ($>$ threshold ) & 32.1 & 26.6 & 23.8 & 22.3 & 22.1 & 21.1 & 20.7 & 19.9 & 19.6 & 20.4 & 20.1 \\
Baseline ($\leq$ threshold ) & 62.4 & 60.3 & 58.6 & 56.9 & 55.8 & 55.0 & 54.3 & 53.9 & 53.6 & 52.1 & 51.9 \\ 
C2RNet ($>$ threshold ) & 33.2 & 28.2 & 25.1 & 23.5 & 23.0 & 22.3 & 22.1 & 21.8 & 21.9 & 20.7 & 20.7 \\
C2RNet ($\leq$ threshold ) & 62.0 & 60.0 & 58.4 & 56.9 & 56.0 & 55.2 & 54.6 & 54.1 & 53.6 & 52.6 & 52.4 \\
\hline
Difference ($>$ threshold )& 1.1 & 1.6 & 1.3 & 1.2 & 0.9 & 1.2 & 1.4 & 1.9 & 2.3 & 0.8 & 0.6 \\ 
Difference ($\leq$ threshold ) & -0.4 & -0.3 & -0.2 & 0.0 & 0.2 & 0.2 & 0.3 & 0.2 & 0.0 & 0.4 & 0.5 \\ 
\hline
\end{tabular}
}
\caption{\label{tbl:detail2}
Rhetorical relation recognition accuracies in predicting text spans with different lengths. \textit{$>$ threshold} represents the overall accuracy when predicting the text spans greater than the given length. \textit{$\leq$ threshold} represents the overall accuracy when predicting the text spans less or equal to the given length. \textit{Difference} represents the gap between the baseline model and our C2RNet. 
}
\end{table*}

In this section, we present a detailed analysis of the performance of the baseline model and our proposed C2RNet model in predicting different text spans. We evaluate their performance on nuclear recognition and rhetorical relation recognition, which are reported in Table \ref{tbl:detail1} and Table \ref{tbl:detail2}, respectively.
Both the baseline model and our C2RNet model employ the fixed T5 language models. The metric \textit{$\leq$ threshold} represents the overall accuracy when predicting text spans that are less than or equal to the specified length threshold. Conversely, the metric \textit{$>$ threshold} represents the overall accuracy when predicting text spans that are greater than the specified length threshold. The \textit{Difference} column quantifies the performance gap between the baseline model and our C2RNet model.

From both tables, a consistent trend is observed: as the length of the text span increases, the accuracies for both nuclear recognition and rhetorical relation recognition tend to decrease. This finding is consistent for both the baseline model and our proposed model, indicating the challenge of accurately recognizing subtrees with relatively larger text spans. 
Our C2RNet demonstrates improved performance compared to the baseline model, particularly for subtrees with moderately long text spans. As presented in Table \ref{tbl:detail1}, when predicting nuclear relations, our model consistently achieves better performance across all span lengths. Notably, for subtrees with lengths greater than $3$, our model surpasses the baseline model by $1.6\%$, and for subtrees with lengths greater than $6$, the margin increases to $3.4\%$. A similar trend is observed in the rhetorical relation recognition results shown in Table \ref{tbl:detail2}. 
The observed enhancement in performance can be attributed to the integration of document-level content knowledge. This integration aids the model in acquiring and processing pivotal information more efficiently.

To provide a simplified overview, we present Table \ref{tbl:ablation_rf} in Section 3.5, where the text spans are divided into three groups with balanced representation, allowing for a clearer comparison of performance differences between our model and the baseline.

\section{Effect on different language model}
\label{sec:appendix_lm}

\begin{table*}[h]
\centering
\scalebox{0.9}{\begin{tabular}{l|c|c|c|c|c|c|c|c}
\hline
& \multicolumn{4}{|c|}{\textbf{Original Parseval}} & \multicolumn{4}{|c}{\textbf{RST Parseval}}\\ 
\hline
\ & \textbf{S} & \textbf{N} & \textbf{R} & \textbf{F} & \textbf{S} & \textbf{N} & \textbf{R} & \textbf{F}  \\
\hline
\ Baseline (RoBERTa)& $ 72.1 $ & $ 61.2 $ & $ 50.8 $ & $ 49.5 $ 
& $ 86.0 $ & $ 73.2 $ & $ 61.1 $ & $ 60.4 $  \\
\ NCRNet (RoBERTa-all)& $ 72.7 $ & $ 62.4 $ & $ 51.8 $ & $ 50.4 $ 
& $ 86.4 $ & $ 74.1 $ & $ 61.7 $ & $ 61.0 $  \\
\ C2RNet (RoBERTa-RST)& $ 75.6 $ & $ 64.3 $ & $ 53.2 $ & $ 51.8 $ 
& $ 87.8 $ & $ 75.8 $ & $ 63.1 $ & $ 62.6 $  \\
\hline
\ Baseline (BERT)& $ 70.2 $ & $ 58.6 $ & $ 48.1 $ & $ 46.8 $ 
& $ 85.1 $ & $ 72.0 $ & $ 59.3 $ & $ 58.6 $  \\
\ C2RNet (BERT-all)& $ 71.4 $ & $ 59.5 $ & $ 49.3 $ & $ 48.1 $ 
& $ 85.7 $ & $ 72.3 $ & $ 60.0 $ & $ 59.3 $  \\
\ C2RNet (BERT-RST)& $ 76.4 $ & $ 64.6 $ & $ 52.8 $ & $ 51.6 $ 
& $ 88.2 $ & $ 75.7 $ & $ 62.6 $ & $ 62.1 $  \\
\hline
\end{tabular}}
\caption{\label{tbl:ablation}
RST discourse parsing results with different language models, on the test set of the RST dataset, using original Parseval and RST Parseval metrics. 
S, N, R, and F represent Span, Nuclearity, Relation, and Full. 
\textit{Baseline (RoBERTa)} and \textit{Baseline (BERT)} represent the baseline models without the NDP branch and using the corresponding language models. 
\textit{C2RNet (RoBERTa-all)} and \textit{C2RNet (BERT-all)} represent C2RNet using RoBERTa or BERT as the shared language model. 
\textit{C2RNet (RoBERTa-RST)} and \textit{C2RNet (BERT-RST)} represents C2RNet using RoBERTa or BERT in RST branch. 
}
\end{table*} % ablation

There are several reasons why we choose to freeze the parameters of pretrained language models. Firstly, our objective is to introduce a high-level content representation to enhance the RST task. By fixing the language model parameters, we can focus on leveraging the content information provided by the NDP branch without introducing additional complexity from joint training on language models. This allows us to assess the performance gain specifically attributable to the content structure-aware representations, rather than the joint learning paradigm. 
Furthermore, freezing the language model parameters enhances the flexibility of our pipeline. If there are other tasks that could potentially benefit RST parsing or other tasks, we can seamlessly integrate their decoders into our system without the need for extensive retraining of language models using data from all these tasks. This saves significant time and effort in adapting the pipeline to new tasks or expanding its capabilities. 
Lastly, by freezing the language model parameters, we maintain the generalization ability of the pretrained models and ensure that the trainable parameters remain lightweight. This helps to reduce computational requirements and facilitates efficient deployment of the system.

To investigate the impact of different language models, we evaluate the performance of C2RNet using RoBERTa \cite{Liu2019RoBERTaAR} and BERT \cite{devlin-etal-2019-bert} (large version) in the RST branch. The results are presented in Table \ref{tbl:ablation}. Specifically, \textit{C2RNet (RoBERTa-all)} and \textit{C2RNet (BERT-all)} denote the models where RoBERTa or BERT serves as the shared fixed language model for both branches. In this configuration, the sentence embeddings obtained from the NDP branch may not be as representative, thus potentially affecting the overall performance of RST. We discuss this issue in the Limitation section.
On the other hand, \textit{C2RNet (RoBERTa-RST)} and \textit{C2RNet (BERT-RST)} represent the models where RoBERTa or BERT is employed in the RST branch, while fixed word embeddings from T5 are used in the NDP branch. 

Although these language models lead to a performance decline, our C2RNet still surpasses the baseline models. This finding suggests that incorporating high-level content-related knowledge benefits RST discourse parsing, particularly in terms of nuclearity and rhetorical relation predictions.

\section{Introduction of News Discourse Profiling }
\label{sec:appendix_ndp}

The NewsDiscourse dataset \cite{choubey-etal-2020-discourse} was created for news discourse profiling, which consists of $802$ news articles. 
Following \citet{choubey-huang-2021-profiling-news}, $502$ documents are used for training, $100$ documents for validation and $200$ documents for testing. 
These data are only used for training the NDP branch. 

The news discourse profiling task aims to analyze the news event structure of news articles. The motivation behind this task lies in the significance of detecting and integrating discourse structures to enhance language comprehension. It categorizes the contents of news articles based on the main event and studies genre-specific discourse structures. Unlike existing tasks that primarily focus on understanding rhetorical aspects or detecting shallow topic transition structures, this task emphasizes the comprehension of global content organization structures with the main event as the central element. 

The content types in the news discourse profiling task revolve around the main news event as the central focus. These content types encompass various aspects, such as the description of the main news event (\textit{Main Event}) and its immediate consequences (\textit{Consequence}). Additionally, there are content types related to providing contextual information, including previous events (\textit{Previous Events}) and the current context (Background)(\textit{Current Context}), which may serve as causes or preconditions for the main event. The further supportive information consists of historical events (\textit{Historical Event}), anecdotal events (\textit{Anecdotal Event}), evaluations from different parties (\textit{Evaluation}), and speculations about the potential impacts of the main event (\textit{Expectation}).

\begin{table*}[h]
\centering
\scalebox{0.9}{\begin{tabular}{l|c|c|c|c|c|c|c|c}
\hline
& \multicolumn{4}{|c|}{\textbf{Original Parseval}} & \multicolumn{4}{|c}{\textbf{RST Parseval}}\\ 
\hline
\ & \textbf{S} & \textbf{N} & \textbf{R} & \textbf{F} & \textbf{S} & \textbf{N} & \textbf{R} & \textbf{F}  \\ \hline
\ Baseline & $ 75.4 $ & $ 64.1 $ & $ 53.6 $ & $ 52.1$ 
& $ 87.7 $ & $ 75.0 $ & $ 63.1 $ & $ 62.3 $  \\
\ C2RNet (one-hot) & $ 75.9 $ & $ 65.2 $ & $ 54.8 $ & $ 53.3 $ 
& $ 88.0 $ & $ 76.3 $ & $ 64.3 $ & $ 63.5 $  \\
\ C2RNet (ours) & $ \bf 76.8 $ & $ \bf 66.2 $ & {$  55.4 $} & {$ \bf 53.8$ }
& {$ \bf 88.4 $} & {$ \bf 76.5 $} & {$ \bf 64.5 $} & {$ \bf 63.8 $}  \\ \hline

\ Human & $ 78.7 $ & $ 66.8 $ & $ 57.1 $ & $ 55.0 $ 
& $ 88.3 $ & $ 77.3 $ & $ 65.4 $ & $ 64.7 $ \\
\hline
\end{tabular}}
\caption{\label{tbl:appendex_rst}
RST discourse parsing results on the test set of the RST dataset, using original Parseval and RST Parseval metrics \cite{marcu-2000-rhetorical}. 
S, N, R, and F represent Span, Nuclearity, Relation, and Full. 
The results of ours are averaged over $3$ random runs and the highest performances are in bold. 
\textit{Baseline} represents the model without an NDP branch and other configurations are kept the same. 
\textit{C2RNet (one-hot)} represents the model where the NDP final one-hot predictions are sent to the RST branch. 
}
\end{table*} % compared with sota

\section{Effect of NDP labels}
\label{sec:appendix_ndpl}

The model represented by the second row, denoted as \textit{C2RNet (one-hot)} in Table \ref{tbl:appendex_rst}, incorporates the final classification layer of the NDP branch and utilizes the NDP one-hot predictions as input for the RST branch. This experiment aims to examine the effectiveness of different forms of information for enhancing RST performance. Comparing \textit{C2RNet (one-hot)} with the \textit{Baseline} model, it is observed that the former outperforms the latter, indicating that the direct utilization of event content labels in the RST parsing task through simple one-hot vector representations is advantageous. However, despite its improved performance, \textit{C2RNet (one-hot)} still exhibits lower performance compared to our final C2RNet, suggesting that the sentence embeddings generated by the NDP branch contain additional essential information that is beneficial for the RST branch.

To further evaluate what information is the main contribution to the RST performance increase, we conducted a probing test towards sentence embeddings generated by the NDP branch in the context of the news discourse profiling task. 
Specifically, we feed the test data of the NDP task to the trained C2RNet and put back the final classification layer of the original NDP model to see how the performance changes. 
And we find its performance merely drops a little. 
These performance drops can be attributed to the fact that the weights of the final prediction layer were originally trained for NDP and remain fixed, while the preceding layers have been updated with RST data. Despite the observed drops, the relatively modest decline in performance suggests that the NDP branch still provides valuable event content information and it is this information that facilitates the RST task.

\end{document}